\documentclass[11pt, a4paper, logo, twocolumn, external, copyright, numbering]{googledeepmind}
\usepackage{xcolor}
\usepackage{tabularx}

\usepackage[authoryear, sort&compress, round]{natbib}
\def\model{Synth$^2$}
\def\dataset{GenPair} 
\definecolor{darkgray}{rgb}{0.43,0.43,0.43}
\title{\model: Boosting Visual-Language Models with Synthetic Captions and Image Embeddings}

\correspondingauthor{sharifzadeh@google.com, abanino@google.com}

\keywords{synthetic data, visual language models, multimodal self-improvement, open-ended learning}
\paperurl{arxiv.org/abs/2403.07750}
\reportnumber{001} 

\author[*,1]{Sahand Sharifzadeh}

\author[1]{Christos Kaplanis}
\author[1]{Shreya Pathak}
\author[1]{Dharshan Kumaran}
\author[1]{Anastasija Ilic}
\author[1]{Jovana Mitrovic}
\author[1]{Charles Blundell}

\author[*,1]{Andrea Banino}
\affil[*]{Equal contributions}
\affil[1]{Google DeepMind}

\begin{abstract}

The creation of high-quality human-labeled image-caption datasets presents a significant bottleneck in the development of Visual-Language Models (VLMs). In this work, we investigate an approach that leverages the strengths of Large Language Models (LLMs) and image generation models to create synthetic image-text pairs for efficient and effective VLM training. Our method employs a pretrained text-to-image model to synthesize image embeddings from captions generated by an LLM. Despite the text-to-image model and VLM initially being trained on the same data, our approach leverages the image generator's ability to create novel compositions, resulting in synthetic image embeddings that expand beyond the limitations of the original dataset.  Extensive experiments demonstrate that our VLM, finetuned on synthetic data achieves comparable performance to models trained solely on human-annotated data, while requiring significantly less data. Furthermore, we perform a set of analyses on captions which reveals that semantic diversity and balance are key aspects for better downstream performance. Finally, we show that synthesizing images in the image embedding space is 25\% faster than in the pixel space. We believe our work not only addresses a significant challenge in VLM training but also opens up promising avenues for the development of self-improving multi-modal models.

\end{abstract}

\begin{document}

\maketitle

\section{Introduction}
\label{submission}

\begin{figure}[t]
  \centering
   \includegraphics[width=1.0\linewidth]{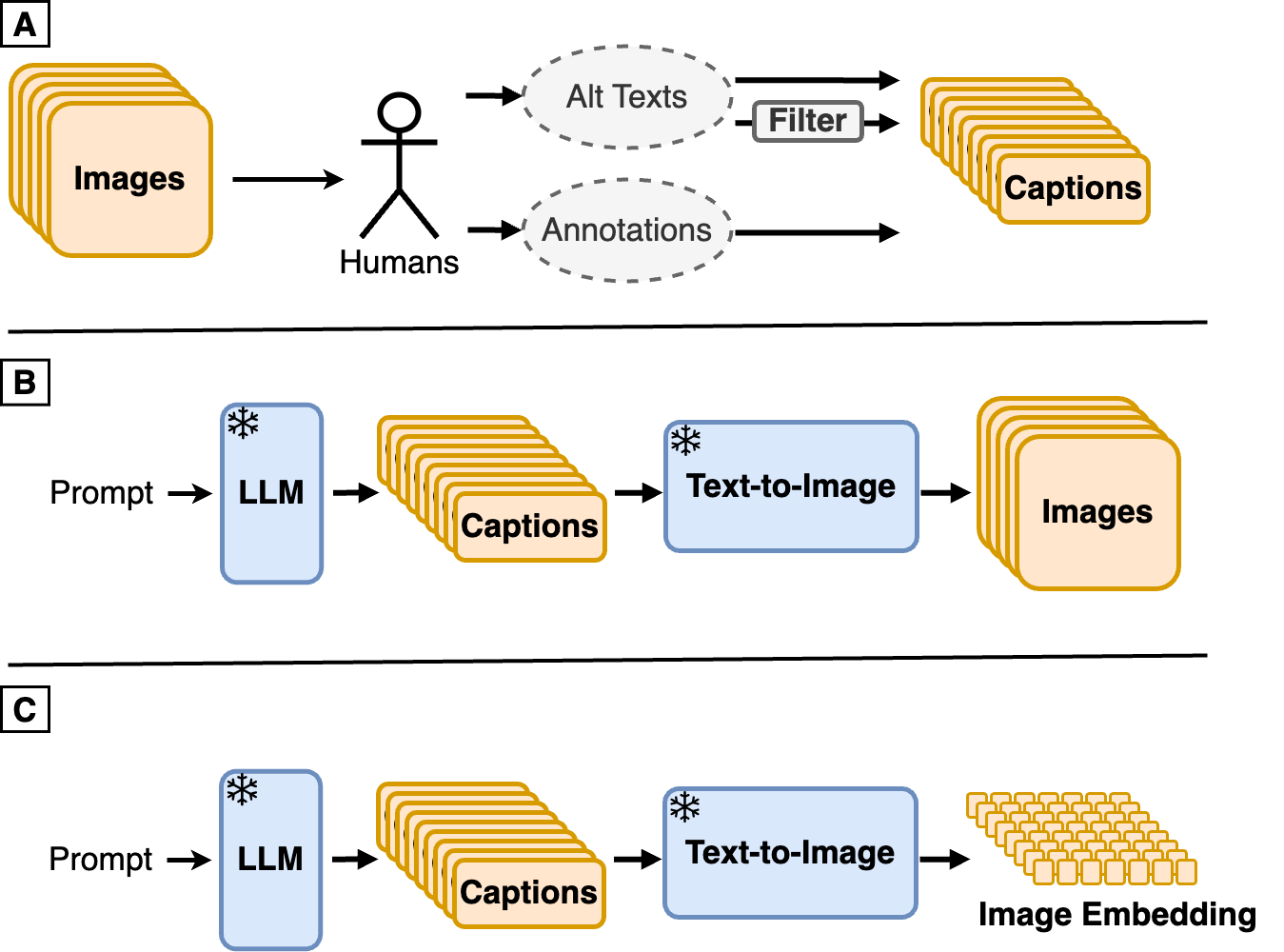}
   \caption{(A) Traditional dataset curation pipelines require a human in the loop to collect and annotate images. (B) We study whether we can reverse this pipeline with generative models, i.e. by first sampling synthetic captions from an LLM and then synthetically generating images from those. (C) By operating in the image embedding space, we also propose to bypass computationally expensive encoder/decoder steps, optimizing and integrating the process within VLM training.}
   \label{fig:onecol}
\end{figure}

Visual-language models (VLMs) are quickly emerging as powerful tools for understanding visual and textual information. Their ability to combine these two modalities holds immense promise for applications ranging from image captioning to visual question answering. While VLMs hold significant potential, their performance is often constrained by limited data availability. Recent breakthroughs demonstrate that pre-training VLMs on larger image-text pair datasets leads to significant improvements in downstream tasks \citep{li2022blip, hu2022scaling}. However, creating such datasets poses several challenges such as scarcity of paired data, potentially noisy nature of data sourced from the internet (e.g., LAION \citep{schuhmann2021laion}), high curation costs, low diversity and high imbalance in semantics. These factors often lead to laborious filtering and extended training times due to low signal-to-noise ratio, thus increasing overall resource consumption (Figure~\ref{fig:onecol}A).

This work tackles these limitations by introducing an efficient approach that leverages pre-trained generative text and image models to create synthetic paired data for VLMs (see Figure~\ref{fig:onecol}B). Our approach uniquely synthesizes both text and images, overcoming the reliance on real-world data and addressing the challenges of scarcity, cost, and noise. 

Additionally, the proposed framework operates at both the pixel and embedding levels (see Figure~\ref{fig:onecol}C), enabling us to train the VLM from either real images or synthetic image embeddings. This bypasses the need for pixel-space rendering for both the image generator and the VLM encoder. This paradigm shift significantly reduces memory overhead and resource consumption while maintaining the quality of synthetic training data.

While synthetic data generation has been explored for various computer vision tasks such as image segmentation, optical flow prediction, or image classification~\citep{mishra2022task2sim,greff2022kubric,azizi2023synthetic, fan2023scaling, li2023data}, its application to both visual and textual modalities within VLM training is a significant advancement. Furthermore, a potential pitfall in studying the impact of synthetic data in training new models lies in the inherent advantage of off-the-shelf image generators. These models, trained on massive image-text datasets, may inadvertently equip VLMs with knowledge already embedded in the image generators' own pre-training data. This overlap could mask the true effectiveness of synthetic images, leading to improvements that might be equally achievable by simply training VLMs on the original image generator dataset.

To isolate the unique contribution of synthetic images, we propose a controlled study. Both the text-to-image generator and the VLM undergo pre-training on the exact same dataset. This alignment ensures that any subsequent improvements observed in the VLM after fine-tuning on synthetic data can be confidently attributed to the novel compositions and representations generated, rather than pre-existing knowledge inherited from the original dataset. 

To sum up, our research explores the synergy between VLMs and text-to-image generation models, demonstrating a powerful framework for overcoming data limitations. This paper presents the following key contributions:

\begin{itemize}
\item \textbf{Fully Synthetic Data Creation:} We introduce the first VLM training process that utilizes a fully synthetic dataset of high-quality text-image embedding pairs. These pairs are generated by pre-trained generative models, circumventing the need for massive amounts of real-world data for either modality.

\item \textbf{Efficient Embedding Space Generation:} Our method works with images in embedding space, avoiding costly image rendering and dramatically improving efficiency, without comprising performance.
\item \textbf{Fair Evaluation Through Control:} By pre-training a text-to-image model on the same dataset used for VLM training, instead of using a large, off-the-shelf model, we prevent knowledge transfer from a model trained on vast, external image-text pairs. This fosters a fair and focused assessment of the benefits of synthetic data in isolation.
\item \textbf{Demonstrated Effectiveness:} Experiments will showcase significant performance gains in 3 domains (scene description, scene understating QA and Extrernal Knowledge QA) when using our synthetic data, assessed against real and synthetic-data baselines.
\end{itemize}

In conclusion, we offer insights into the future of VLM training, highlighting the potential for creating customized datasets, open-ended training and self-improving multimodal models.

\section{Related Work}

\renewcommand{\arraystretch}{1.1}
\begin{table*}[t]
\centering
\caption{A taxonomy of related work on synthetic data for training vision models.}
\label{taxonomy}
\resizebox{1.0\textwidth}{!}{\begin{tabular}{l|l|c|c|c|c|cc}
\toprule
Generator & Method & Generator Model & Generated Set & Caption Class & Caption Type & Evaluation Setting
 \\
 \midrule
Canonical Concept Mapping
 & \cite{sharifzadeh2021classification} & Linear &  Scene Graph Embedding & Scene Graphs &  &  \\
  & \cite{sharifzadeh2022improving} &  & & Complex Text & Human Generated & SG Classification\\
  \cline{1-8}

  & \citet{mishra2022task2sim} & & Mixed pairs such as & &  & \\ & \citet{greff2022kubric} & & (Segmentation, Images) & &  & & \\ Simulation/Rendering Engine &\citet{zheng2020structured3d} & Mix &  (Optimal Flow, Videos)  & N/A & N/A & Mix & \\ & \citet{de2022next} & & (Depth Maps, Images) & & & \\
  \cline{2-8}
 & \citet{cascante2023going} & Mix &  (Captions, Images) & Complex Text & Rule-based & Vision Encoder \\

\cline{1-8}
 & \citet{azizi2023synthetic} & SD  & &  &   & 
  \\
  Off-the-shelf Image Generator & \citet{fan2023scaling} & Imagen &  Images & Single Word & ImageNet Classes & Classifier
  \\ 
  &  & MUSE &  & &  &  &  \\
  \cline{2-8}
 & \citet{li2023data} & SD & Images & Complex Text & Human Generated & VLM  \\

\cline{1-8}
\textbf{Controlled Image Generator}
 & \textbf{\model} & \textbf{MUSE} & \textbf{Text, Embeddings \& Images} & \textbf{Complex Text} & \textbf{Human} \textbf{\&} \textbf{LLM Generated} & \textbf{VLM} \\

\bottomrule
\end{tabular}}
\end{table*}

Visual-language models (VLMs) pretrained on large-scale image-text data, primarily sourced from the web, have shown remarkable capabilities across numerous vision-language tasks. These tasks include image captioning, visual question answering, and few-shot learning~\citep{alayrac2022flamingo, chen2022pali, li2022blip}. Architecturally, VLMs commonly employ an image encoder coupled with a large language model (LLM). Researchers explore diverse image encoder designs, ranging from convolutional neural networks (CNNs)~\citep{alayrac2022flamingo} to transformer-based architectures~\citep{chen2022pali}. Additionally, the choice between pretraining the image encoder on separate image datasets~\citep{alayrac2022flamingo} and training it from scratch alongside the VLM~\citep{tsimpoukelli2021multimodal} remains an active area of investigation. Pretraining VLMs involves various objectives, including supervised learning approaches, as well as contrastive methods that focus either on aligning image and text representations~\citep{radford2021learning} or on aligning different image representations~\citep{chen2020big,chen2020simple,grill2020bootstrap}. Similar diversity exists in LLM choices, with both encoder-decoder~\citep{chen2022pali} and decoder-only~\citep{alayrac2022flamingo} architectures being utilized. The strategies for combining information extracted by the image encoder and the language model also exhibit substantial variation. Despite the success achieved with web-scraped datasets, there's a growing emphasis on designing VLM architectures that facilitate training using synthetically generated data. This capability holds particular significance for applications where data scarcity or resources availability pose significant challenges.

Generation of synthetic data for training machine learning models remains a highly active area of research. While numerous studies~\citep{mishra2022task2sim,cascante2023going,greff2022kubric, zheng2020structured3d,de2022next} explore the use of model-based rendering engines or simulators, the remarkable advancements in high-quality image generators have ignited a surge of interest in leveraging generative models for creating training data. This trend has significantly impacted a wide range of computer vision tasks, including semantic segmentation~\citep{li2022bigdatasetgan,ros2016synthia,baranchuk2021label,tritrong2021repurposing,li2021semantic,chen2019learning}, human motion understanding~\citep{varol2017learning,ma2022pretrained,guo2022learning}, and more recently image classification~\citep{azizi2023synthetic, fan2023scaling}. Our work investigates the potential of data-driven generative models within the domain of visual-language models (VLMs). We focus on their application in downstream tasks, where the ability to process complex scene is important. From this aspect, our approach is closest to a concurrent work by~\citet{li2023data} where the goal is to replace faulty images in a captioning pipeline with their synthetic versions. However, our work distinguishes itself from~\citet{li2023data} by achieving strong performance while utilizing 40 times less paired data and only a quarter of the parameters. This demonstrates the potential for achieving both performance and efficiency through our proposed approach. Furthermore, unlike the mentioned works \citep{azizi2023synthetic, fan2023scaling,li2023data}, we avoid using an off-the-self image generator trained on larger datasets. This approach prevented knowledge transfer into the VLM from a model trained on large amounts of paired data, which is critical as such knowledge transfer would have obscured our ability to scientifically assess the unique contribution of synthetic data.

Our work can also be likened to the concept of cycle consistency~\citep{zhu2017unpaired} in visual-language models. This principle, where image-to-text and text-to-image conversion can be employed as a form of synthetic data training, albeit with extra gradients throughout the network, exploring image-text cycle consistency during sampling~\citep{li2023dall} or training~\citep{li2023leveraging} has been explored in recent works,  demonstrating promising results.

Finally, we emphasize the efficiency of pipelines that seamlessly connect data generation and model training. While most prior work has concentrated on generating images in pixel space, we investigate the generation of image embeddings that can be directly integrated into the VLM. Our approach aligns with recent work in the scene graph classification domain~\citep{sharifzadeh2021classification,sharifzadeh2022improving}, which has explored synthetic scene graph embedding generation using pretrained canonical class representations. 

Table~\ref{taxonomy} provides a taxonomy of some of the mentioned related work to clarify the differences better. As shown, our study is the first of its kind to scientifically explore the application of synthetic data generation within a VLM training pipeline and in particular using image embeddings and by generating synthetic captions.

\label{sec:approach}
\section{\model}
Given the synthetic generation of text and then images, we refer to our method as \model. \model \space is a pipeline for training VLMs using generative models in addition to collected human data. In this section we introduce different components of this pipeline, namely Caption Generation, Image Generation and the full \model \space model (see Figure~\ref{fig:archs} for details).

\subsection{Synthetic Caption Generation}\label{captions}

\begin{figure}
    \centering
    \includegraphics[width=1.0\linewidth]{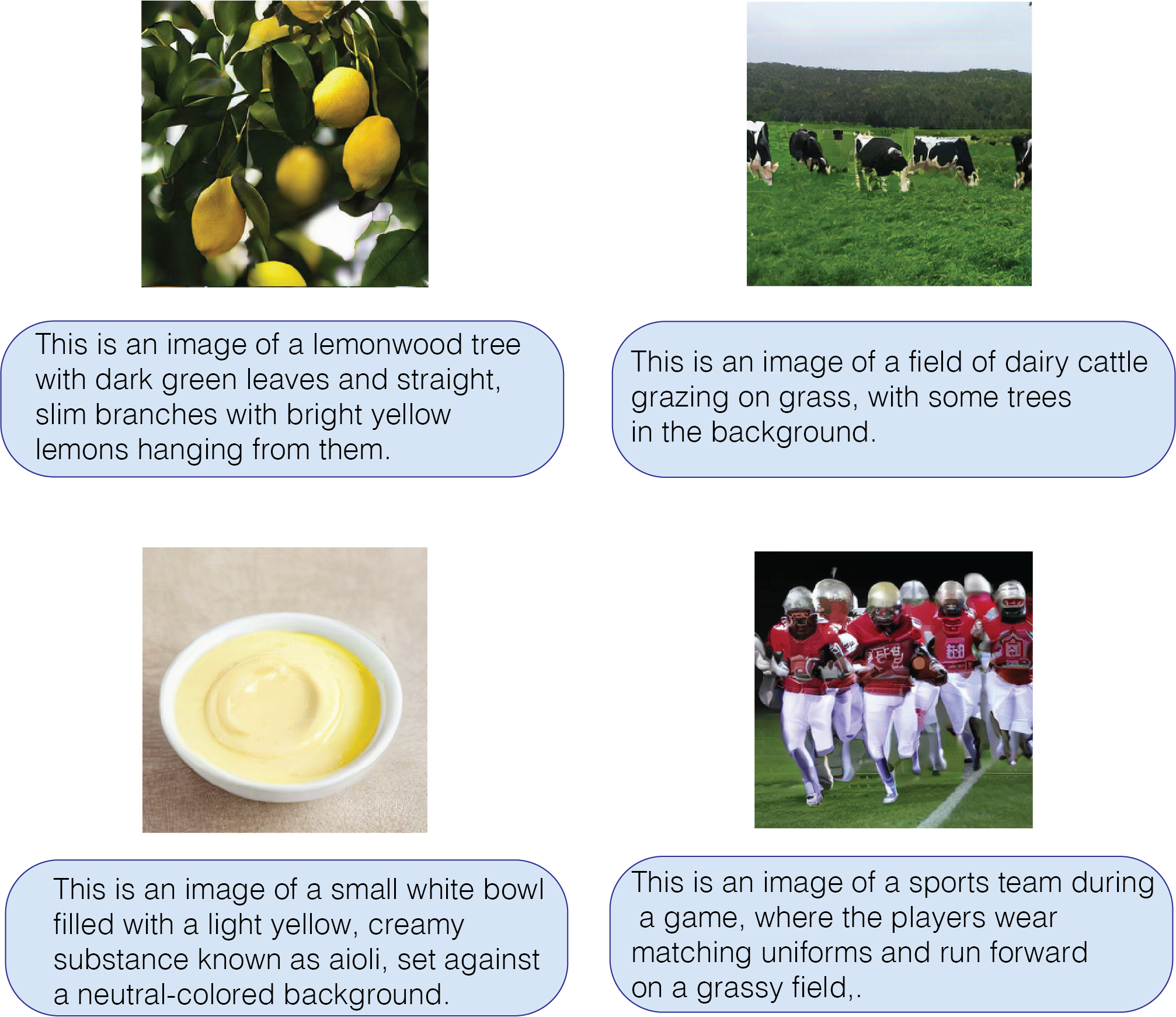}
    \caption{Examples of synthetic captions and synthetic images generated by LLM and text-to-image generator.}
    \label{fig:captions}
\end{figure}

We leverage the generative capabilities of LLMs to create synthetic captions. To ensure a wide variety of captions, we adopt a class-based prompting approach. First, we randomly select a class from the ImageNet21k dataset \citep{ridnik2021imagenet}. The LLM (Gemini Pro~\citep{team2023gemini}) is then prompted with the following text:

\textit{Make up a human-annotated description of an image that contains the following object: $[object]$. The caption should be around 30-40 words long. Describe the different components of the scene in an objective and unbiased way. Do not add subjective judgments about the image, it should be as factual as possible. Do not use fluffy, poetic language. Respond only with the caption itself, beginning with ``This is an image of''.}

where we replace ``[object]'' with the randomly selected class. This class-based prompting encourages the LLM to generate captions that cover a broad range of visual concepts. Figure~\ref{fig:captions} shows some samples from the generated captions.

\subsection{Image Generation}\label{t2i}


Our image generator architecture is similar to~\citet{chang2023muse} and was chosen for its superior inference efficiency due to parallel decoding and its use of discrete image tokens that dramatically reduces the number of sampling iterations compared to auto-regressive models. Additionally, MUSE leverages a transformer architecture, further enhancing efficiency \citep{chang2023muse}. As discussed in Section~\ref{submission}, to mitigate the potential for knowledge overlap between off-the-shelf image generators and VLMs, we deviate from standard practice. Instead of employing pre-trained image generators, we intentionally pre-train both our text-to-image generator and VLM on an identical dataset. This controlled setup ensures that any performance gains observed in the VLM after fine-tuning on the synthetic data are directly attributable to the unique contributions of the generated images, rather than inherited knowledge from the original dataset. Hence, we pretrain an image generator on a image-caption dataset that will also be utilized for VLM pretraining (i.e. Conceptual Captions v2~\citep{changpinyo2021conceptual} see Section \ref{sec:experiments}).

\subsubsection{Training}
To pretrain our image generator given paired image and texts, we embed the texts with a pre-trained language model and the images with a pre-trained VQ-GAN~\citep{esser2021taming} (refer to \ref{sec:vq-gan-details} in the appendix for the details). The pre-trained language model is the same used in our VLM and it is a reduced version of Chinchilla~\citep{hoffmann2022training} using 400 million parameters.
Following the approach in \citet{chang2023muse}, we apply a masking procedure with a cosine scheduling function. This function gradually increases the masking ratio during training, biasing the model towards learning from images with larger masked regions. On average, around 64\% of the VQ tokens are replaced with a special ``dropped token''. The noisy VQ tokens and the embedded text are then fed into a transformer model. The model's objective is to predict the original, unmasked VQ tokens. The loss function is the cross-entropy between the predicted VQ tokens and the masked ones:

$$L_{t2i} = \prod_{t \in M} p(z(x)_t | y, \{z(x)_u | u \notin M\}),$$

where $z(x)$ denotes the VQ tokens computed from the ground-truth image $x$, $y$ denotes the ground-truth caption, and $M$ denotes the indices of the masked tokens. Our text-to-image generator was trained only on 10.1 millions text-image pairs from Conceptual Captions V2 \citep{changpinyo2021conceptual} (see Section \ref{sec:experiments} for details).

\begin{figure}[t]
  \centering
    \includegraphics[width=1.0\linewidth]{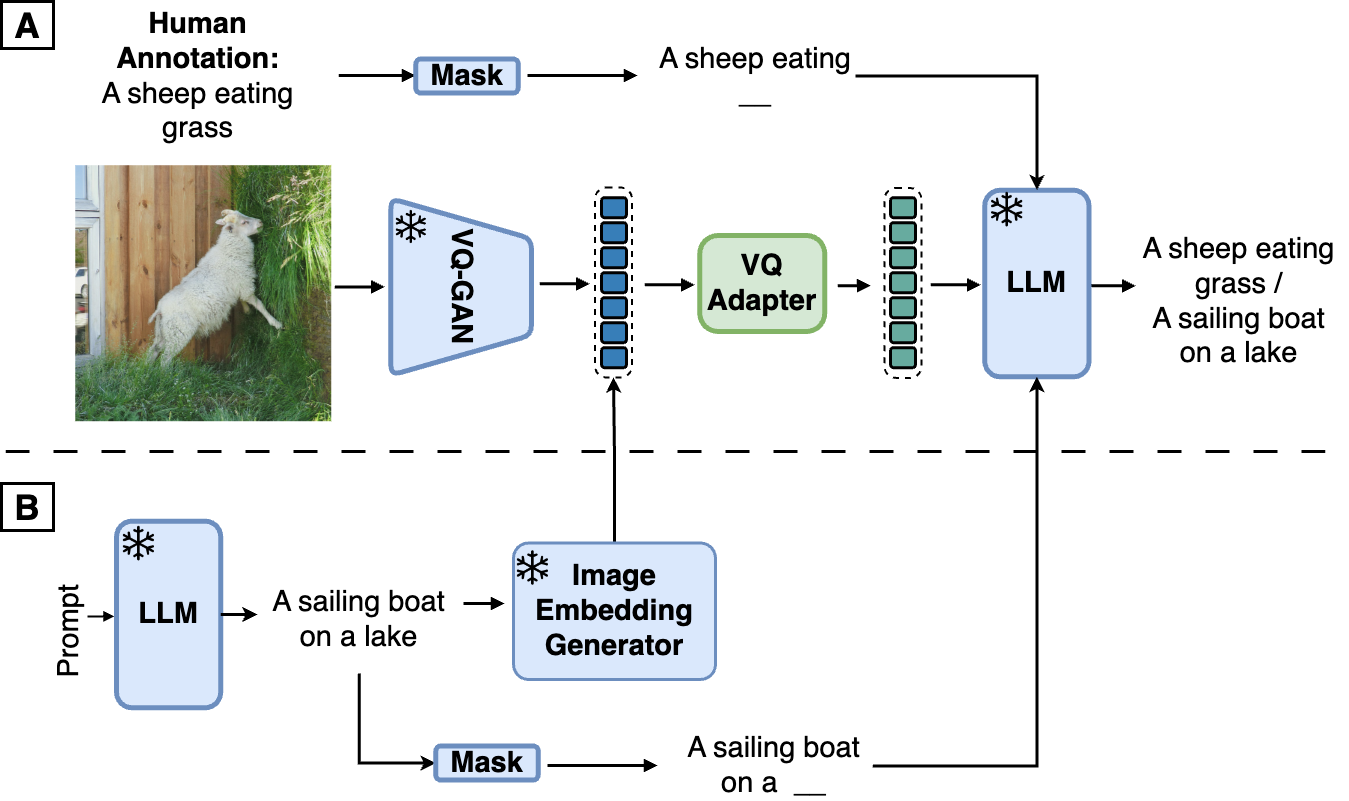}
    \caption{We introduce a VLM framework that leverages LLMs and image generation models to create synthetic image-text pairs for efficient training. We can train a VLM from both non-synthetic (A) and synthetic (B) data as shown in this figure. Our model trained with the added synthetic pairs demonstrates impressive image captioning performance, significantly reducing the need for human annotated images.}
    \label{fig:archs}
\end{figure}

\subsubsection{Inference}
During inference, we use an iterative procedure to generate images. At first iteration, the predicted VQ tokens are all initialized with the ``dropped token'' representing the masked entries. At each upcoming iteration, the text embedding and the predicted VQ tokens until that stage are fed into the transformer and the model predicts the masked tokens. A cosine schedule determines which masked tokens have the highest prediction confidence during decoding. These tokens are unmasked, progressively reducing the masked token set. This process repeats for 24 iterations, gradually refining the generated VQ-tokens. Additional details for our text-to-image generator model and examples of generated images, given ground truth annotations compared to the original ones, are reported in \ref{sec:t2i-details} in the appendix.

\subsection{\model \space VLM architecture}

As described in Section~\ref{t2i}, our text-to-image model generates image embeddings through an iterative caption-conditioned denoising process. A VQ-GAN decoder could be used to convert these tokens into human-viewable images. However, we emphasize that the image embeddings themselves provide a compressed representation of an image, making the decoder step unnecessary for certain applications.

Therefore, to enhance efficiency for training the VLM from synthetic data, we design the VLM to enable the bypass of pixel-space processing. This is achieved by setting its vision encoder to be identical to the VQ-GAN backbone used for our image generator. This enables seamless interaction between the VLM and synthetically generated image embeddings and eliminates the computationally expensive decoding stage of VQ-GAN for the image generator and the encoding stage for the VLM when training on synthetic data. At the same time, it still allows us to train from images in pixel space when using human annotated data.  This design feature is particularly important since in order to curtail model collapse effects~\citep{shumailov2023curse}, we need to mix synthetic and real data during finetuning~\citep{gerstgrasser2024model}.

Overall, our VQ-based design removes the need for costly conversions to and from pixel space, resulting in significant time and disk space savings as will be shown in Section~\ref{sec:experiments}. To harness the efficiency of discrete tokens during training and sampling, while maintaining the rich image understanding provided by soft embeddings, we convert each discrete token to its soft embedding using the codebook before feeding them to the VLM.

On top of the VQ backbone in our VLM, we incorporate a Perceiver Resampler component \citep{jaegle2021perceiver} to cross-attend to the VQ tokens extracted from our backbone. Similar to Flamingo~\cite{alayrac2022flamingo}, we employ a frozen language model to cross-attend to the image representation and autoregressively generate text. While the self-attention layers in the language model remain frozen, the cross-attention layers between the language and vision components are trainable, allowing the model to learn effective multimodal representations.

The VLM's objective is to predict the next token in a sequence, conditioned on the input image. This is formulated as a cross-entropy loss:
\vspace{-8pt} 

\begin{equation}L_{VLM} = \prod_{l=1}^L p(y_l|y_{<l}, x), \end{equation}

where \(x\) denotes the image, \(y_l\) represents the l-th language token, and \(y_{<l}\) represents all tokens preceding \(y_l\). As shown in Figure~\ref{fig:archs} by combining the components introduced in the previous parts, we can train VLMs either from human annotated data or synthetic image-captions pairs such that the VLM is conditioned on the synthetic image embedding with the cross-entropy loss being between the synthetic captions and the VLM's predictions.
\section{Experiments}
\label{sec:experiments}

\subsection{Experimental setup}

\subsubsection{Datasets}
For training we used four datasets: 
(1) \textbf{Conceptual Captions v2}~\citep{changpinyo2021conceptual} (CCv2) is a dataset with 12 million image-text pairs that are collected by automatically extracting images and their corresponding ``alt-tex'' descriptions from various websites. In our experiments we use CCv2 as the main source of human-annotated data whereas for the other datasets we might use synthetically generated images/texts depending on the experiment. 
(2) \textbf{WebLI}~\citep{chen2022pali} has around 350 million image and alt text pairs with high text quality in English.
(3) \textbf{LTIP}~\citep{alayrac2022flamingo} has 312 million web collected image and text pairs.
(4) \textbf{\dataset} \space refers to 1 million fully synthetic captions paired with synthetic images generated that we create on-the-fly during the VLM training using our text-to-image generator. The details are described in Section~\ref{captions}.

We tested our model on 3 different domains:
\begin{itemize}
    \item \textbf{Scene description}: for this task we used MS-COCO~\citep{chen2015microsoft} and we evaluate the performance of our models on the commonly used test set of MS-COCO under zero-shot and fine-tuning settings. For the finetuning settings we use the training set MS-COCO. We use the Karpathy split~\citep{karpathy2015deep} for evaluation. Another dataset we used for this class is Flickr-30k~\citep{plummer2015flickr30k}: again, we evaluate performances on the Karpathy split.  We use CIDEr score~\citep{vedantam2015cider} as our metric.
    \item \textbf{Scene understanding QA}: for this task we used VQAv2~\citep{goyal2017making} and we evaluated on the test-dev set. We use VQA accuracy \citep{antol2015vqa} as our metric.
    \item \textbf{External knowledge QA}: for this task we used OKVQA~\citep{marino2019ok} and we evaluated on the validations set as in \citet{alayrac2022flamingo}. We use VQA accuracy \citep{antol2015vqa} as our metric.
\end{itemize}

\subsubsection{Training details} For both the image generator and the VLM we use a pre-trained and frozen VQ-GAN (see \ref{sec:vq-gan-details} for details on the network). The images are input at a resolution of 256x256, with a patch size of 16, which results in a sequence length of 256 tokens (16x16). Both models use a pre-trained and frozen Chinchilla 400m~\citep{hoffmann2022training} as LLM. The VLM has a perceiver resampler~\citep{alayrac2022flamingo, jaegle2021perceiver} with 6 layers and 64 latent input queries that cross-attend to the image embeddings. There are cross attention layers between each layer in the LLM and the output of perceiver resampler similar to \citep{alayrac2022flamingo} ( refer to \ref{sec:perceiver-details} for details). 

We use ViT-B~\citep{dosovitskiy2020image} for the main text-to-image model in the image generator (see Appendix~\ref{sec:t2i-details} for full details), with a maximum text length of 64 tokens with guidance scale of 4.0, 24 refining steps and a sample choice temperature of 32.5. All the models are trained with AdamW~\citep{loshchilov2017decoupled} with a learning rate of 0.0001, 5000 warmup steps, and the batch size of 512. For finetuning settings (on COCO), we use a learning rate of 0.00001. Section \ref{sec:t2i-details} reports additional details. We pre-train the image generator for 500k steps at which point it has a Fréchet inception distance (FID)~\citep{heusel2017gans} of 17.1 on MS-COCO test pairs. Our VLMs training experiments all run for 500k steps. Our models are all implemented in JAX and trained on 256 TPUs.

\subsection{Results}

\begin{table*}[t]
\centering
\caption{Zero shot image captioning results when training with ground truth captions paired with either the original or synthetically generated images.}
\vspace{0.3cm}
\label{tab:results-compensate-real-cap}
\resizebox{1.0\textwidth}{!}{\begin{tabular}{l|lcccccccc}
\toprule
Method & Real & Synth & \#Real Data & \#Synth Data & CIDEr-COCO ($\uparrow$)  & CIDEr-Flickr-30 ($\uparrow$) & VQAV2 Accuracy ($\uparrow$) & OKVQA Accuracy ($\uparrow$)\\
\midrule
Baseline & CCv2 & - & 10.1M & - & 22.1 & 12.7 & 29.1 & 32.4 \\
\midrule
\model & CCv2 & LTIP & 10.1M & 330M & 28.7 & 16.7 & 33.5 & 34.5 \\
& CCv2 & WebLI & 10.1M & 350M & 27.6 & 18.1 & 34.1 & 34.7 \\
& CCv2 & WebLI+LTIP & 10.1M & 670M & \textbf{33.4} & \textbf{20.8} & \textbf{35.3} & \textbf{36.1} \\
\midrule
\textcolor{darkgray}{
Gold} & \textcolor{darkgray}{CCv2+LTIP} & - & \textcolor{darkgray}{340.1M} & - & \textcolor{darkgray}{27.6} & \textcolor{darkgray}{20.9} & \textcolor{darkgray}{36.7} & \textcolor{darkgray}{38.1} \\
& \textcolor{darkgray}{CCv2+WebLI} & - & \textcolor{darkgray}{360.1M} & - & \textcolor{darkgray}{30.7} & \textcolor{darkgray}{21.1} & \textcolor{darkgray}{36.2} & \textcolor{darkgray}{38.5} \\
& \textcolor{darkgray}{CCv2+LTIP+WebLI} & - & \textcolor{darkgray}{690.1M} & - & \textcolor{darkgray}{35.3} & \textcolor{darkgray}{23.4} & \textcolor{darkgray}{37.9} & \textcolor{darkgray}{39.4} \\

\bottomrule
\end{tabular}}
\end{table*}

\begin{table*}[t]
\centering
\caption{Zero shot results when using synthetically generated caption and image embedding pairs. Concentration is calculated as the cumulative distribution on the top-5 clusters, a lower value represent higher diversity (see Appendix~\ref{sec:app-cluster} for more details).} 
\vspace{0.2cm}
\label{tab:results-pure-synth}
\resizebox{1.0\textwidth}{!}{\begin{tabular}{lcccccccccc}
\toprule
Real & Synth & \#Real Data & \#Synth Data & Concentration ($\downarrow$) & Entropy ($\uparrow$) & CIDEr-COCO ($\uparrow$) & Flickr-30($\uparrow$) & VQAV2 Acc. ($\uparrow$) & OKVQA Acc. ($\uparrow$)\\
\midrule
CCv2 & - & 10.1M & -  & - & - & 22.1 & 12.7 & 29.1 & 32.4\\
CCv2 & \dataset & 10.1M & 1M  & \textbf{57.7}\%  & \textbf{3.81} & \textbf{25.9} & \textbf{17.3} &\textbf{31.1} & \textbf{34.0}\\
CCv2+WebLI & -  & 10.1M+1M & -  & 69.8\% & 3.43 & 24.4 & 14.9 & 30.6 & \textbf{33.9}\\
CCv2+LTIP &  - & 10.1M+1M & - & 83.0\% & 2.92 & 23.4 & 13.8 & 30.3 & 32.9\\

\bottomrule
\end{tabular}}
\end{table*}

\subsubsection{Synthetic Images}
\label{human-captions-exp}

To first assess the effectiveness of synthetic images compared to the original ones in VLM training, we conduct a study where human-written captions from existing datasets are paired with synthetic images generated by a text-to-image model. We train a VLM on these synthetic image-caption pairs and evaluate its performance against a model trained on the original real image-caption pairs. This comparison allows us to investigate whether synthetic images can effectively substitute for real images in the context of VLM training.

Specifically, we generate synthetic images for the LTIP and WebLI datasets using their annotations. This provides us with \textit{(Real Text, Synthetic Image)} pairs to train a VLM with (\textbf{\model} in Table~\ref{tab:results-compensate-real-cap}). We compare this model to the model trained on 
the human-annotated data of Conceptual Caption v2 (CCv2) only, using no data from LTIP and WebLI (\textbf{Baseline}) and the models trained on
the original \textit{(Real Text, Real Image)} pairs (\textbf{Gold} in Table~\ref{tab:results-compensate-real-cap}). For consistency, and to reduce model collapse, all \model \space and Gold models are co-trained with CCv2, the same dataset used by our text-to-image generator and the baseline.

As shown in Table~\ref{tab:results-compensate-real-cap}, synthetic images significantly improve baseline model performance across the full set of downstream tasks.  Importantly, they are effective for  VLM training despite the smaller volume of human-annotated images used. Note that ``Gold'' represents the upper performance bound when using the full original dataset. For LTIP, synthetic images slightly outperform original images, likely due to increased data diversity; while original pairs remain static during training, synthetic images introduce beneficial image augmentation~\citep{azizi2023synthetic}.

These results suggest the potential of generative models to enhance VLM training containing complex, multi-object scenes and detailed captions, even with limited real image data. 

\subsubsection{Synthetic text and image pairs}
\label{synt-captions-exp}

To further demonstrate the efficacy of using synthetic data for VLM training, we conducted additional experiments where the entire dataset including captions and their corresponding images are generated synthetically by an LLM and the image generator (see Section \ref{sec:approach} for the explanation of how these caption where generate and Figure~\ref{fig:onecol}D for some examples). 

Comparing the first two rows in Table~\ref{tab:results-pure-synth} shows that adding even a small fraction (1M) of purely synthetic image-caption data (\dataset) significantly improves performance across all evaluated domains. Interestingly, sampling an additional 1M data points from \textit{real} datasets like WebLI or LTIP (rows 3 and 4) yields lower improvement.

To investigate this, we assessed the semantic diversity and balance of each caption set. First, we used our language model to generate embeddings for the synthetic GenPair and the real WebLI and LTIP captions. Then, we employed k-means clustering to analyze these embeddings (see Appendix \ref{sec:app-cluster} for details). Figure~\ref{fig:diversity} shows the cluster distribution among all three datasets, with the x-axis representing the cluster index and the y-axis representing  data volume from each dataset within a cluster. Notice how LTIP and WebLI have a large concentration of data only within clusters 19, 18, 1, 16 and 0, while most other clusters are under-populated. In contrast, GenPair is distributed more evenly across different clusters (e.g. clusters 1, 4, 6, 9, 12, 14, 15, 16, 17, 18 are all well populated), suggesting superior balance compared to the other two. 

To quantify the mentioned semantic concentration within each dataset we use a ``Concentration Metric'' as the  percentage of data belonging to the top-5 most populated clusters. As reported in Table~\ref{tab:results-pure-synth}, GenPair has the lowest concentration, with only 57.5\% of its captions in the top-5 clusters. This indicates lower semantic imbalance in the synthetically generated GenPair and contrasts with the higher imbalance found in the other datasets (69.9\% and 83\%). Also, the entropy over the full cluster distribution, confirms that GenPair has a more uniform distribution of data across clusters. We postulate that the inherent diversity within GenPair likely contributes to its robust performance on downstream tasks, as models trained on diverse data tend to generalize better. 

This evaluation demonstrates \model's capacity to leverage diverse text data, including synthetic data, for enhanced performance. Our proposed analysis and metrics can inform synthetic data creation, further optimizing its impact. Notably, our findings highlight the significant role of synthetic data, generated via image synthesis and LLM captions, in overcoming data limitations and boosting VLM capabilities. This underscores the effectiveness of text-only captions for model enhancement, emphasizing the value of our technique.

\begin{figure}[t]
  \centering
    \includegraphics[width=1.0\linewidth]{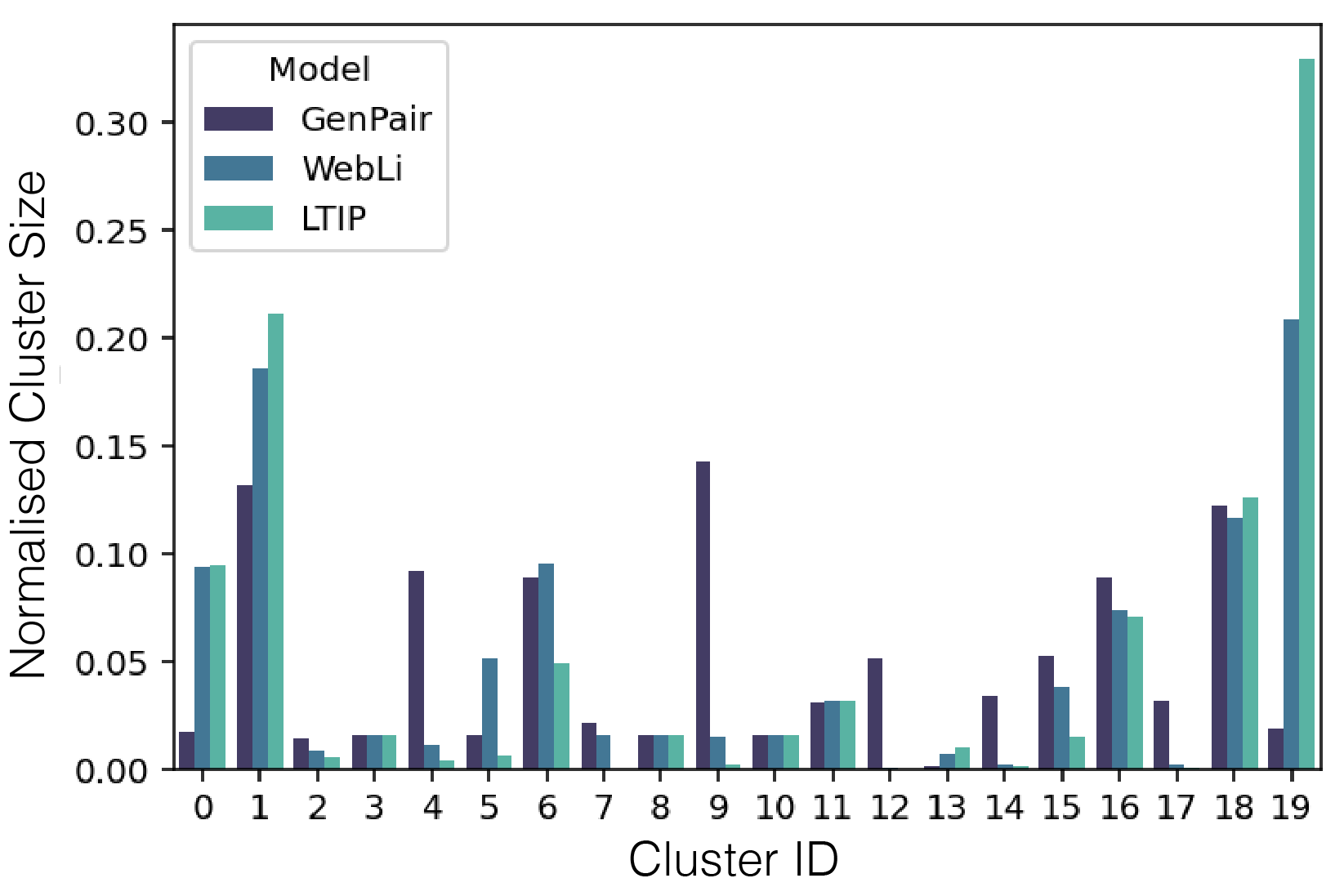}
    \vspace*{-6mm}
    \caption{Semantic diversity. Histogram represent the distribution of cluster sizes, with GenPair showing a more uniform coverage of semantic concepts. See \ref{sec:app-cluster} for details on how the histogram was derived.} 
    \label{fig:diversity}
\end{figure}

\begin{figure}[t]
  \centering
    \includegraphics[width=1.0\linewidth]{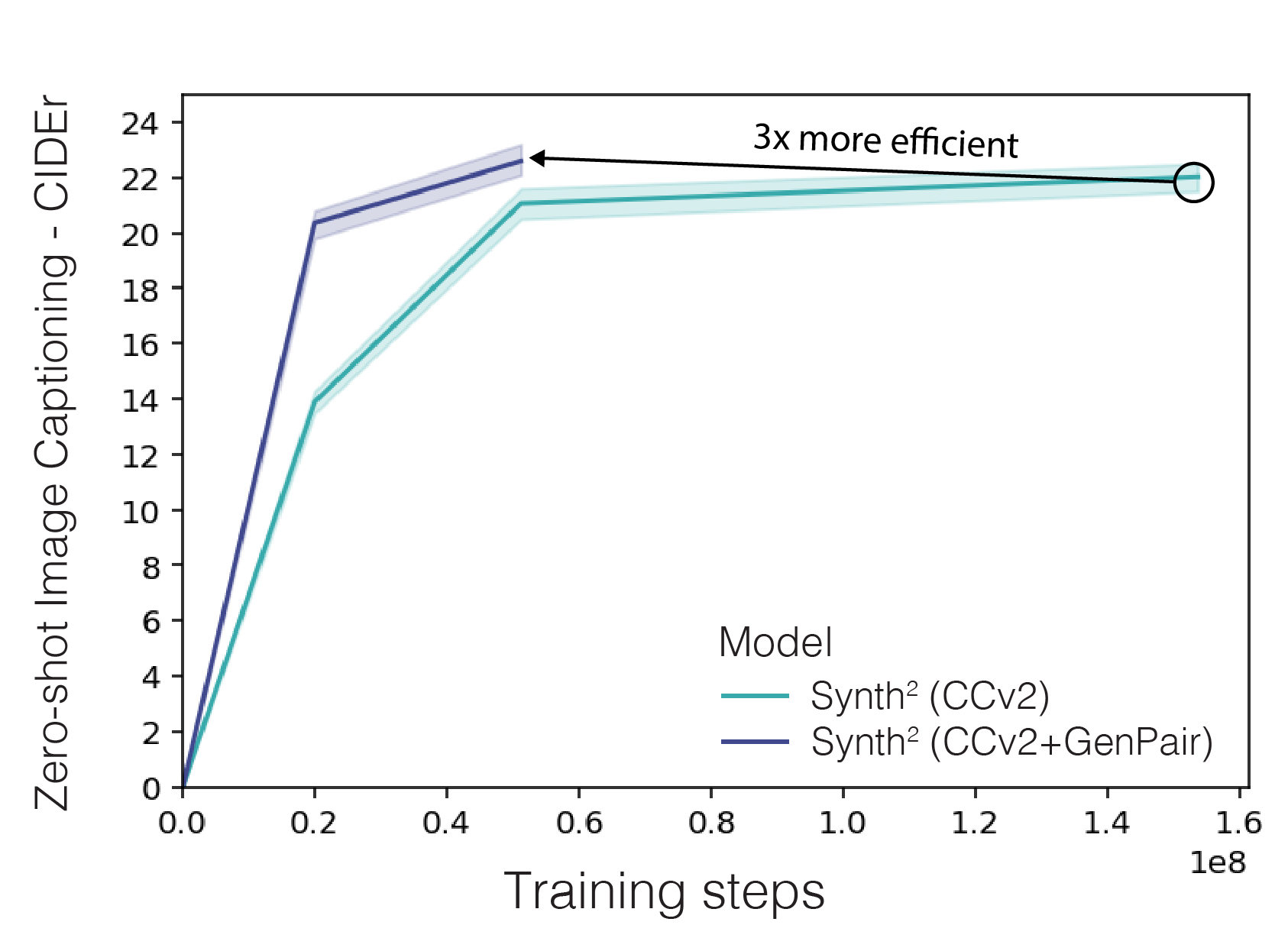}
    \vspace*{-6mm}
    \caption{ Performance as a function of training steps. The blue curve shows the baseline trained solely on paired data (CCv2). The purple curve demonstrates \model's performance trained additionally on augmentation with fully synthetic data (\dataset). \model \space achieves parity with the baseline using roughly 1/3 of the training steps, showcasing its superior efficiency. Shaded regions represent standard deviation across 3 random seeds.}
    \label{fig:scale}
\end{figure}



\subsubsection{Comparison with related work}

Furthermore, we investigated the potential of our approach compared to the state-of-the-art model while taking into account the amount of synthetic or real training data and the number of model parameters. We combined all the three datasets namely, \dataset \space, WebLI and LTIP. Table~\ref{tab:comparison} compares our model's MS-COCO benchmark performance against related methods: ITIT~\citep{li2023leveraging}, DC~\citep{li2023data} and SimVLM~\citep{wang2021simvlm}. Note that, in this case we provide both zero-shot and finetuning results (see Section~\ref{sec:app-fine-tuning} in the Appendix). ITIT employs cycle consistency for VLM improvement, while DC uses Stable Diffusion to generate images based on human-defined heuristics, and SimVLM has one of the state-of-the-art performances on MS-COCO.

\model \space and ITIT share similar parameter counts and paired data usage, making them directly comparable. As shown in Table~\ref{tab:comparison}, \model \space drastically outperforms ITIT on the CIDEr score. While DC achieves the highest raw performance, it requires significantly more parameters, relies on vast amount of real data, and the filtering procedure is heavily based on human heuristics.

\model \space strikes a balance between performance and efficiency. By leveraging text-only data, it achieves competitive results using 40 times less paired data than DC. Interestingly, despite performing weaker than SimVLM on finetuning,  \model \space has the highest in zero-shot performance among all models. This comparison highlights the trade-offs between performance, efficiency, and data requirements. \model \space emerges as a compelling solution when resources are limited, achieving impressive results with reduced data and computational overhead. 

\subsubsection{Extra analysis}

To further analyze the boost in performance afforded by augmentation with fully synthetic data, we characterized the performance of the baseline model (i.e. trained solely on CCv2), and \model \space which was additionally trained together with \dataset. Figure~\ref{fig:scale} visualizes the performance trends for both models. Notably, the light blue curve representing the baseline exhibits a gradual increase in performance with increasing training steps, eventually plateauing at 22.1 CIDER score. In contrast, the purple curve reveals \model's steeper performance improvement, achieving comparable performance to the paired only training regime, with roughly 1/3 of the training steps. This highlights the significant data efficiency gain achieved through fully synthetic data augmentation in \model. Shaded regions surrounding each curve indicate standard deviation across three random seeds, demonstrating the robustness of the observed performance trends. These findings demonstrate \model's ability to effectively leverage fully synthetic data, resulting in remarkable performance gains.

Our study also investigated the computational efficiency of \model. We compared the case where captions are used to generate image embedding versus when captions are rendered into actual images. As shown in Table~\ref{tab:efficiency}, \model \space trained using image embedding consistently demonstrates faster performance, running at 2.08 training steps per second, whereas \model \space trained from pixel runs at a slower pace of 1.66 steps/second. Critically, the performance on image captioning and visual question answering are not affected. This efficiency advantage stems from \model's utilization of text-only data and embedding-based image generation using parallel decoding, which reduces the computational overhead associated with processing pixel-based images as done in the \model \space model trained from pixels without affecting the performance. For further studies, e.g. comparing the performance of different image encoder backbones refer to Appendix, Table~\ref{tab:backbone}.

\begin{table}[t]
\centering
\caption{Zero shot and fine tuning results, comparison with related work. The image-captioning performance is evaluated on the COCO Karpathy split. Bold indicates the best and underline the second best values.}
\vspace{0.2cm}
\label{tab:comparison}
\resizebox{0.5\textwidth}{!}{\begin{tabular}{lcccccc}
\toprule
Model & $\#$All Params$\downarrow$ & $\#$Real Data$\downarrow$ & $\#$Synth Data$\downarrow$ & CIDEr$\uparrow$ (FT) &
CIDEr$\uparrow$ (zs)\\
\midrule
ITIT & 11.2B & \textbf{3M} & \textbf{110M} & 103.5 & 32.1  \\
DC (BLIP) & 1.7B & 5.5B & 400M & \underline{133.1} & -  \\
SimVLM & 1.4B & 1.1B & 365M &  \textbf{143.3} & 32.2  \\
\model & \textbf{632M} & \underline{10.1M} & 711M & 131.3 & \textbf{35.4}  \\

\bottomrule
\end{tabular}}
\end{table}

\begin{table}[t]
\centering
\caption{Model efficiency measured by computing the steps per second of training on the same hardware. \model \space in embedding space consistently outperforms the \model \space using pixels, demonstrating superior training efficiency.}
\vspace{0.2cm}
\label{tab:efficiency}
\resizebox{0.5\textwidth}{!}{\begin{tabular}{lcccc}
\toprule
Model & Step/Sec$\uparrow$ & MSCOCO CIDEr(zs)$\uparrow$ &
VAQV2 Accuracy (ft)$\uparrow$\\
\midrule
\model - Embeddings & \textbf{2.08} & 25.9 & 50.1\\
\model - Pixels & 1.66 & 26.0 & 49.9\\
\bottomrule
\end{tabular}}
\end{table}

\section{Limitations}
\label{limitations}
The two main limitations of the current work stem from the restricted quantity of fully synthetic data used and the limited exploration of text data sources. Firstly, whilst we were able to show a substantial gain when training was augmented with fully synthetic data, our experiments were limited to a relatively low quantity of data (i.e. 1M image-caption pairs). In the future, it will be important to examine whether using orders of magnitude more fully synthetic data (e.g. $\sim$700M) will result in performance gains that surpass that of using solely paired data. 
Also, the current work primarily explores a restricted set of text-only data sources. Further research is necessary to investigate the potential of diverse text sources, including investigating context-specific generation (e.g., medical data), which could be an exciting avenue for future exploration. By acknowledging and addressing these limitations, future research can leverage the strengths of the \model \space approach and enhance its generalizability, robustness, and efficiency for a broader range of visual-language modeling tasks.

\section{Conclusions}

Our study presents a novel approach for generating synthetic image-text pairs to enhance the training of visual language models. By harnessing the capabilities of large language models and text-to-image generation, we effectively address the limitations of manual image labeling such as scalability, customization and costs. In particular, we show that the visual language model trained on our synthetic and human annotated datasets exhibits a marked improvement in image captioning tasks compared to a baseline trained exclusively on human-annotated data. Overall, this research introduces a novel technique that has the potential to transform visual language model training. Our findings highlight the value of synthetic data generation, paving the way for advancements in numerous fields where visual language understanding is crucial.

\section{Broader Impact}
\label{broader-impact}
This paper presents work whose goal is to advance the field of Machine Learning. There are many potential societal consequences of our work, none which we feel must be specifically highlighted here. In particular biases in generative models can have a drastic effect on the synthetic data. Similarly it is important to consider the challenges of privacy when using generative models.

\section*{Acknowledgements}

We would like to thank Sander Dieleman, Ali Razavi, and Benigno Uria for their insights on VQGAN. We would also like to thank Aida Nematzadeh, Pinelopi Papalampidi, Han Zhang, Mateusz Malinowski, Valentin De Bortoli, Sahra Ghalebikesabi, Emanuele Bugliarello, Chris Knutsen, and Murray Shanahan for their in depth comments and support throughout the project.

\bibliography{main}

\begin{thebibliography}{54}
\providecommand{\natexlab}[1]{#1}
\providecommand{\url}[1]{\texttt{#1}}
\expandafter\ifx\csname urlstyle\endcsname\relax
  \providecommand{\doi}[1]{doi: #1}\else
  \providecommand{\doi}{doi: \begingroup \urlstyle{rm}\Url}\fi

\bibitem[Alayrac et~al.(2022)Alayrac, Donahue, Luc, Miech, Barr, Hasson, Lenc,
  Mensch, Millican, Reynolds, et~al.]{alayrac2022flamingo}
J.-B. Alayrac, J.~Donahue, P.~Luc, A.~Miech, I.~Barr, Y.~Hasson, K.~Lenc,
  A.~Mensch, K.~Millican, M.~Reynolds, et~al.
\newblock Flamingo: a visual language model for few-shot learning.
\newblock \emph{Advances in Neural Information Processing Systems},
  35:\penalty0 23716--23736, 2022.

\bibitem[Antol et~al.(2015)Antol, Agrawal, Lu, Mitchell, Batra, Zitnick, and
  Parikh]{antol2015vqa}
S.~Antol, A.~Agrawal, J.~Lu, M.~Mitchell, D.~Batra, C.~L. Zitnick, and
  D.~Parikh.
\newblock Vqa: Visual question answering.
\newblock In \emph{Proceedings of the IEEE international conference on computer
  vision}, pages 2425--2433, 2015.

\bibitem[Azizi et~al.(2023)Azizi, Kornblith, Saharia, Norouzi, and
  Fleet]{azizi2023synthetic}
S.~Azizi, S.~Kornblith, C.~Saharia, M.~Norouzi, and D.~J. Fleet.
\newblock Synthetic data from diffusion models improves imagenet
  classification.
\newblock \emph{arXiv preprint arXiv:2304.08466}, 2023.

\bibitem[Baranchuk et~al.(2021)Baranchuk, Rubachev, Voynov, Khrulkov, and
  Babenko]{baranchuk2021label}
D.~Baranchuk, I.~Rubachev, A.~Voynov, V.~Khrulkov, and A.~Babenko.
\newblock Label-efficient semantic segmentation with diffusion models.
\newblock \emph{arXiv preprint arXiv:2112.03126}, 2021.

\bibitem[Brock et~al.(2021)Brock, De, Smith, and Simonyan]{brock2021high}
A.~Brock, S.~De, S.~L. Smith, and K.~Simonyan.
\newblock High-performance large-scale image recognition without normalization.
\newblock In \emph{International Conference on Machine Learning}, pages
  1059--1071. PMLR, 2021.

\bibitem[Cascante-Bonilla et~al.(2023)Cascante-Bonilla, Shehada, Smith, Doveh,
  Kim, Panda, Varol, Oliva, Ordonez, Feris, et~al.]{cascante2023going}
P.~Cascante-Bonilla, K.~Shehada, J.~S. Smith, S.~Doveh, D.~Kim, R.~Panda,
  G.~Varol, A.~Oliva, V.~Ordonez, R.~Feris, et~al.
\newblock Going beyond nouns with vision \& language models using synthetic
  data.
\newblock In \emph{Proceedings of the IEEE/CVF International Conference on
  Computer Vision}, pages 20155--20165, 2023.

\bibitem[Chang et~al.(2023)Chang, Zhang, Barber, Maschinot, Lezama, Jiang,
  Yang, Murphy, Freeman, Rubinstein, et~al.]{chang2023muse}
H.~Chang, H.~Zhang, J.~Barber, A.~Maschinot, J.~Lezama, L.~Jiang, M.-H. Yang,
  K.~Murphy, W.~T. Freeman, M.~Rubinstein, et~al.
\newblock Muse: Text-to-image generation via masked generative transformers.
\newblock \emph{arXiv preprint arXiv:2301.00704}, 2023.

\bibitem[Changpinyo et~al.(2021)Changpinyo, Sharma, Ding, and
  Soricut]{changpinyo2021conceptual}
S.~Changpinyo, P.~Sharma, N.~Ding, and R.~Soricut.
\newblock Conceptual 12m: Pushing web-scale image-text pre-training to
  recognize long-tail visual concepts.
\newblock In \emph{Proceedings of the IEEE/CVF Conference on Computer Vision
  and Pattern Recognition}, pages 3558--3568, 2021.

\bibitem[Chen et~al.(2020{\natexlab{a}})Chen, Kornblith, Norouzi, and
  Hinton]{chen2020simple}
T.~Chen, S.~Kornblith, M.~Norouzi, and G.~Hinton.
\newblock A simple framework for contrastive learning of visual
  representations.
\newblock In \emph{International conference on machine learning}, pages
  1597--1607. PMLR, 2020{\natexlab{a}}.

\bibitem[Chen et~al.(2020{\natexlab{b}})Chen, Kornblith, Swersky, Norouzi, and
  Hinton]{chen2020big}
T.~Chen, S.~Kornblith, K.~Swersky, M.~Norouzi, and G.~E. Hinton.
\newblock Big self-supervised models are strong semi-supervised learners.
\newblock \emph{Advances in neural information processing systems},
  33:\penalty0 22243--22255, 2020{\natexlab{b}}.

\bibitem[Chen et~al.(2015)Chen, Fang, Lin, Vedantam, Gupta, Doll{\'a}r, and
  Zitnick]{chen2015microsoft}
X.~Chen, H.~Fang, T.-Y. Lin, R.~Vedantam, S.~Gupta, P.~Doll{\'a}r, and C.~L.
  Zitnick.
\newblock Microsoft coco captions: Data collection and evaluation server.
\newblock \emph{arXiv preprint arXiv:1504.00325}, 2015.

\bibitem[Chen et~al.(2022)Chen, Wang, Changpinyo, Piergiovanni, Padlewski,
  Salz, Goodman, Grycner, Mustafa, Beyer, et~al.]{chen2022pali}
X.~Chen, X.~Wang, S.~Changpinyo, A.~Piergiovanni, P.~Padlewski, D.~Salz,
  S.~Goodman, A.~Grycner, B.~Mustafa, L.~Beyer, et~al.
\newblock Pali: A jointly-scaled multilingual language-image model.
\newblock \emph{arXiv preprint arXiv:2209.06794}, 2022.

\bibitem[Chen et~al.(2019)Chen, Li, Chen, and Gool]{chen2019learning}
Y.~Chen, W.~Li, X.~Chen, and L.~V. Gool.
\newblock Learning semantic segmentation from synthetic data: A geometrically
  guided input-output adaptation approach.
\newblock In \emph{Proceedings of the IEEE/CVF conference on computer vision
  and pattern recognition}, pages 1841--1850, 2019.

\bibitem[de~Melo et~al.(2022)de~Melo, Torralba, Guibas, DiCarlo, Chellappa, and
  Hodgins]{de2022next}
C.~M. de~Melo, A.~Torralba, L.~Guibas, J.~DiCarlo, R.~Chellappa, and
  J.~Hodgins.
\newblock Next-generation deep learning based on simulators and synthetic data.
\newblock \emph{Trends in cognitive sciences}, 2022.

\bibitem[Dosovitskiy et~al.(2020)Dosovitskiy, Beyer, Kolesnikov, Weissenborn,
  Zhai, Unterthiner, Dehghani, Minderer, Heigold, Gelly,
  et~al.]{dosovitskiy2020image}
A.~Dosovitskiy, L.~Beyer, A.~Kolesnikov, D.~Weissenborn, X.~Zhai,
  T.~Unterthiner, M.~Dehghani, M.~Minderer, G.~Heigold, S.~Gelly, et~al.
\newblock An image is worth 16x16 words: Transformers for image recognition at
  scale.
\newblock \emph{arXiv preprint arXiv:2010.11929}, 2020.

\bibitem[Esser et~al.(2021)Esser, Rombach, and Ommer]{esser2021taming}
P.~Esser, R.~Rombach, and B.~Ommer.
\newblock Taming transformers for high-resolution image synthesis.
\newblock In \emph{Proceedings of the IEEE/CVF conference on computer vision
  and pattern recognition}, pages 12873--12883, 2021.

\bibitem[Fan et~al.(2023)Fan, Chen, Krishnan, Katabi, Isola, and
  Tian]{fan2023scaling}
L.~Fan, K.~Chen, D.~Krishnan, D.~Katabi, P.~Isola, and Y.~Tian.
\newblock Scaling laws of synthetic images for model training... for now.
\newblock \emph{arXiv preprint arXiv:2312.04567}, 2023.

\bibitem[Gerstgrasser et~al.(2024)Gerstgrasser, Schaeffer, Dey, Rafailov,
  Sleight, Hughes, Korbak, Agrawal, Pai, Gromov, et~al.]{gerstgrasser2024model}
M.~Gerstgrasser, R.~Schaeffer, A.~Dey, R.~Rafailov, H.~Sleight, J.~Hughes,
  T.~Korbak, R.~Agrawal, D.~Pai, A.~Gromov, et~al.
\newblock Is model collapse inevitable? breaking the curse of recursion by
  accumulating real and synthetic data.
\newblock \emph{arXiv preprint arXiv:2404.01413}, 2024.

\bibitem[Goyal et~al.(2017)Goyal, Khot, Summers-Stay, Batra, and
  Parikh]{goyal2017making}
Y.~Goyal, T.~Khot, D.~Summers-Stay, D.~Batra, and D.~Parikh.
\newblock Making the v in vqa matter: Elevating the role of image understanding
  in visual question answering.
\newblock In \emph{Proceedings of the IEEE conference on computer vision and
  pattern recognition}, pages 6904--6913, 2017.

\bibitem[Greff et~al.(2022)Greff, Belletti, Beyer, Doersch, Du, Duckworth,
  Fleet, Gnanapragasam, Golemo, Herrmann, et~al.]{greff2022kubric}
K.~Greff, F.~Belletti, L.~Beyer, C.~Doersch, Y.~Du, D.~Duckworth, D.~J. Fleet,
  D.~Gnanapragasam, F.~Golemo, C.~Herrmann, et~al.
\newblock Kubric: A scalable dataset generator.
\newblock In \emph{Proceedings of the IEEE/CVF Conference on Computer Vision
  and Pattern Recognition}, pages 3749--3761, 2022.

\bibitem[Grill et~al.(2020)Grill, Strub, Altch{\'e}, Tallec, Richemond,
  Buchatskaya, Doersch, Avila~Pires, Guo, Gheshlaghi~Azar,
  et~al.]{grill2020bootstrap}
J.-B. Grill, F.~Strub, F.~Altch{\'e}, C.~Tallec, P.~Richemond, E.~Buchatskaya,
  C.~Doersch, B.~Avila~Pires, Z.~Guo, M.~Gheshlaghi~Azar, et~al.
\newblock Bootstrap your own latent-a new approach to self-supervised learning.
\newblock \emph{Advances in neural information processing systems},
  33:\penalty0 21271--21284, 2020.

\bibitem[Guo et~al.(2022)Guo, Wu, Wang, Su, Su, Gan, Huang, and
  Yang]{guo2022learning}
X.~Guo, W.~Wu, D.~Wang, J.~Su, H.~Su, W.~Gan, J.~Huang, and Q.~Yang.
\newblock Learning video representations of human motion from synthetic data.
\newblock In \emph{Proceedings of the IEEE/CVF Conference on Computer Vision
  and Pattern Recognition}, pages 20197--20207, 2022.

\bibitem[Heusel et~al.(2017)Heusel, Ramsauer, Unterthiner, Nessler, and
  Hochreiter]{heusel2017gans}
M.~Heusel, H.~Ramsauer, T.~Unterthiner, B.~Nessler, and S.~Hochreiter.
\newblock Gans trained by a two time-scale update rule converge to a local nash
  equilibrium.
\newblock \emph{Advances in neural information processing systems}, 30, 2017.

\bibitem[Hoffmann et~al.(2022)Hoffmann, Borgeaud, Mensch, Buchatskaya, Cai,
  Rutherford, Casas, Hendricks, Welbl, Clark, et~al.]{hoffmann2022training}
J.~Hoffmann, S.~Borgeaud, A.~Mensch, E.~Buchatskaya, T.~Cai, E.~Rutherford,
  D.~d.~L. Casas, L.~A. Hendricks, J.~Welbl, A.~Clark, et~al.
\newblock Training compute-optimal large language models.
\newblock \emph{arXiv preprint arXiv:2203.15556}, 2022.

\bibitem[Hu et~al.(2022)Hu, Gan, Wang, Yang, Liu, Lu, and Wang]{hu2022scaling}
X.~Hu, Z.~Gan, J.~Wang, Z.~Yang, Z.~Liu, Y.~Lu, and L.~Wang.
\newblock Scaling up vision-language pre-training for image captioning.
\newblock In \emph{Proceedings of the IEEE/CVF conference on computer vision
  and pattern recognition}, pages 17980--17989, 2022.

\bibitem[Jaegle et~al.(2021)Jaegle, Gimeno, Brock, Vinyals, Zisserman, and
  Carreira]{jaegle2021perceiver}
A.~Jaegle, F.~Gimeno, A.~Brock, O.~Vinyals, A.~Zisserman, and J.~Carreira.
\newblock Perceiver: General perception with iterative attention.
\newblock In \emph{International conference on machine learning}, pages
  4651--4664. PMLR, 2021.

\bibitem[Karpathy and Fei-Fei(2015)]{karpathy2015deep}
A.~Karpathy and L.~Fei-Fei.
\newblock Deep visual-semantic alignments for generating image descriptions.
\newblock In \emph{Proceedings of the IEEE conference on computer vision and
  pattern recognition}, pages 3128--3137, 2015.

\bibitem[Kingma and Ba(2014)]{kingma2014adam}
D.~P. Kingma and J.~Ba.
\newblock Adam: A method for stochastic optimization.
\newblock \emph{arXiv preprint arXiv:1412.6980}, 2014.

\bibitem[Li et~al.(2021)Li, Yang, Kreis, Torralba, and Fidler]{li2021semantic}
D.~Li, J.~Yang, K.~Kreis, A.~Torralba, and S.~Fidler.
\newblock Semantic segmentation with generative models: Semi-supervised
  learning and strong out-of-domain generalization.
\newblock In \emph{Proceedings of the IEEE/CVF Conference on Computer Vision
  and Pattern Recognition}, pages 8300--8311, 2021.

\bibitem[Li et~al.(2022{\natexlab{a}})Li, Ling, Kim, Kreis, Fidler, and
  Torralba]{li2022bigdatasetgan}
D.~Li, H.~Ling, S.~W. Kim, K.~Kreis, S.~Fidler, and A.~Torralba.
\newblock Bigdatasetgan: Synthesizing imagenet with pixel-wise annotations.
\newblock In \emph{Proceedings of the IEEE/CVF Conference on Computer Vision
  and Pattern Recognition}, pages 21330--21340, 2022{\natexlab{a}}.

\bibitem[Li et~al.(2023{\natexlab{a}})Li, Gu, Koner, Sharifzadeh, and
  Tresp]{li2023dall}
H.~Li, J.~Gu, R.~Koner, S.~Sharifzadeh, and V.~Tresp.
\newblock Do dall-e and flamingo understand each other?
\newblock In \emph{Proceedings of the IEEE/CVF International Conference on
  Computer Vision}, pages 1999--2010, 2023{\natexlab{a}}.

\bibitem[Li et~al.(2022{\natexlab{b}})Li, Li, Xiong, and Hoi]{li2022blip}
J.~Li, D.~Li, C.~Xiong, and S.~Hoi.
\newblock Blip: Bootstrapping language-image pre-training for unified
  vision-language understanding and generation.
\newblock In \emph{International Conference on Machine Learning}, pages
  12888--12900. PMLR, 2022{\natexlab{b}}.

\bibitem[Li et~al.(2023{\natexlab{b}})Li, Bhardwaj, Tian, Zhang, Barber,
  Katabi, Lajoie, Chang, and Krishnan]{li2023leveraging}
T.~Li, S.~Bhardwaj, Y.~Tian, H.~Zhang, J.~Barber, D.~Katabi, G.~Lajoie,
  H.~Chang, and D.~Krishnan.
\newblock Leveraging unpaired data for vision-language generative models via
  cycle consistency.
\newblock \emph{arXiv preprint arXiv:2310.03734}, 2023{\natexlab{b}}.

\bibitem[Li et~al.(2023{\natexlab{c}})Li, Lotz, Qiu, and Elliott]{li2023data}
W.~Li, J.~F. Lotz, C.~Qiu, and D.~Elliott.
\newblock Data curation for image captioning with text-to-image generative
  models.
\newblock \emph{arXiv preprint arXiv:2305.03610}, 2023{\natexlab{c}}.

\bibitem[Loshchilov and Hutter(2017)]{loshchilov2017decoupled}
I.~Loshchilov and F.~Hutter.
\newblock Decoupled weight decay regularization.
\newblock \emph{arXiv preprint arXiv:1711.05101}, 2017.

\bibitem[Ma et~al.(2022)Ma, Bai, and Zhou]{ma2022pretrained}
J.~Ma, S.~Bai, and C.~Zhou.
\newblock Pretrained diffusion models for unified human motion synthesis.
\newblock \emph{arXiv preprint arXiv:2212.02837}, 2022.

\bibitem[Marino et~al.(2019)Marino, Rastegari, Farhadi, and
  Mottaghi]{marino2019ok}
K.~Marino, M.~Rastegari, A.~Farhadi, and R.~Mottaghi.
\newblock Ok-vqa: A visual question answering benchmark requiring external
  knowledge.
\newblock In \emph{Proceedings of the IEEE/cvf conference on computer vision
  and pattern recognition}, pages 3195--3204, 2019.

\bibitem[Mishra et~al.(2022)Mishra, Panda, Phoo, Chen, Karlinsky, Saenko,
  Saligrama, and Feris]{mishra2022task2sim}
S.~Mishra, R.~Panda, C.~P. Phoo, C.-F.~R. Chen, L.~Karlinsky, K.~Saenko,
  V.~Saligrama, and R.~S. Feris.
\newblock Task2sim: Towards effective pre-training and transfer from synthetic
  data.
\newblock In \emph{Proceedings of the IEEE/CVF Conference on Computer Vision
  and Pattern Recognition}, pages 9194--9204, 2022.

\bibitem[Plummer et~al.(2015)Plummer, Wang, Cervantes, Caicedo, Hockenmaier,
  and Lazebnik]{plummer2015flickr30k}
B.~A. Plummer, L.~Wang, C.~M. Cervantes, J.~C. Caicedo, J.~Hockenmaier, and
  S.~Lazebnik.
\newblock Flickr30k entities: Collecting region-to-phrase correspondences for
  richer image-to-sentence models.
\newblock In \emph{Proceedings of the IEEE international conference on computer
  vision}, pages 2641--2649, 2015.

\bibitem[Radford et~al.(2021)Radford, Kim, Hallacy, Ramesh, Goh, Agarwal,
  Sastry, Askell, Mishkin, Clark, et~al.]{radford2021learning}
A.~Radford, J.~W. Kim, C.~Hallacy, A.~Ramesh, G.~Goh, S.~Agarwal, G.~Sastry,
  A.~Askell, P.~Mishkin, J.~Clark, et~al.
\newblock Learning transferable visual models from natural language
  supervision.
\newblock In \emph{International conference on machine learning}, pages
  8748--8763. PMLR, 2021.

\bibitem[Ridnik et~al.(2021)Ridnik, Ben-Baruch, Noy, and
  Zelnik-Manor]{ridnik2021imagenet}
T.~Ridnik, E.~Ben-Baruch, A.~Noy, and L.~Zelnik-Manor.
\newblock Imagenet-21k pretraining for the masses.
\newblock \emph{arXiv preprint arXiv:2104.10972}, 2021.

\bibitem[Ros et~al.(2016)Ros, Sellart, Materzynska, Vazquez, and
  Lopez]{ros2016synthia}
G.~Ros, L.~Sellart, J.~Materzynska, D.~Vazquez, and A.~M. Lopez.
\newblock The synthia dataset: A large collection of synthetic images for
  semantic segmentation of urban scenes.
\newblock In \emph{Proceedings of the IEEE conference on computer vision and
  pattern recognition}, pages 3234--3243, 2016.

\bibitem[Schuhmann et~al.(2021)Schuhmann, Vencu, Beaumont, Kaczmarczyk, Mullis,
  Katta, Coombes, Jitsev, and Komatsuzaki]{schuhmann2021laion}
C.~Schuhmann, R.~Vencu, R.~Beaumont, R.~Kaczmarczyk, C.~Mullis, A.~Katta,
  T.~Coombes, J.~Jitsev, and A.~Komatsuzaki.
\newblock Laion-400m: Open dataset of clip-filtered 400 million image-text
  pairs.
\newblock \emph{arXiv preprint arXiv:2111.02114}, 2021.

\bibitem[Sharifzadeh et~al.(2021)Sharifzadeh, Baharlou, and
  Tresp]{sharifzadeh2021classification}
S.~Sharifzadeh, S.~M. Baharlou, and V.~Tresp.
\newblock Classification by attention: Scene graph classification with prior
  knowledge.
\newblock In \emph{Proceedings of the AAAI Conference on Artificial
  Intelligence}, volume~35, pages 5025--5033, 2021.

\bibitem[Sharifzadeh et~al.(2022)Sharifzadeh, Baharlou, Schmitt, Sch{\"u}tze,
  and Tresp]{sharifzadeh2022improving}
S.~Sharifzadeh, S.~M. Baharlou, M.~Schmitt, H.~Sch{\"u}tze, and V.~Tresp.
\newblock Improving scene graph classification by exploiting knowledge from
  texts.
\newblock In \emph{Proceedings of the AAAI Conference on Artificial
  Intelligence}, volume~36, pages 2189--2197, 2022.

\bibitem[Shumailov et~al.(2023)Shumailov, Shumaylov, Zhao, Gal, Papernot, and
  Anderson]{shumailov2023curse}
I.~Shumailov, Z.~Shumaylov, Y.~Zhao, Y.~Gal, N.~Papernot, and R.~Anderson.
\newblock The curse of recursion: Training on generated data makes models
  forget.
\newblock \emph{arXiv preprint arXiv:2305.17493}, 2023.

\bibitem[Team et~al.(2023)Team, Anil, Borgeaud, Wu, Alayrac, Yu, Soricut,
  Schalkwyk, Dai, Hauth, et~al.]{team2023gemini}
G.~Team, R.~Anil, S.~Borgeaud, Y.~Wu, J.-B. Alayrac, J.~Yu, R.~Soricut,
  J.~Schalkwyk, A.~M. Dai, A.~Hauth, et~al.
\newblock Gemini: a family of highly capable multimodal models.
\newblock \emph{arXiv preprint arXiv:2312.11805}, 2023.

\bibitem[Tritrong et~al.(2021)Tritrong, Rewatbowornwong, and
  Suwajanakorn]{tritrong2021repurposing}
N.~Tritrong, P.~Rewatbowornwong, and S.~Suwajanakorn.
\newblock Repurposing gans for one-shot semantic part segmentation.
\newblock In \emph{Proceedings of the IEEE/CVF conference on computer vision
  and pattern recognition}, pages 4475--4485, 2021.

\bibitem[Tsimpoukelli et~al.(2021)Tsimpoukelli, Menick, Cabi, Eslami, Vinyals,
  and Hill]{tsimpoukelli2021multimodal}
M.~Tsimpoukelli, J.~L. Menick, S.~Cabi, S.~Eslami, O.~Vinyals, and F.~Hill.
\newblock Multimodal few-shot learning with frozen language models.
\newblock \emph{Advances in Neural Information Processing Systems},
  34:\penalty0 200--212, 2021.

\bibitem[Varol et~al.(2017)Varol, Romero, Martin, Mahmood, Black, Laptev, and
  Schmid]{varol2017learning}
G.~Varol, J.~Romero, X.~Martin, N.~Mahmood, M.~J. Black, I.~Laptev, and
  C.~Schmid.
\newblock Learning from synthetic humans.
\newblock In \emph{Proceedings of the IEEE conference on computer vision and
  pattern recognition}, pages 109--117, 2017.

\bibitem[Vedantam et~al.(2015)Vedantam, Lawrence~Zitnick, and
  Parikh]{vedantam2015cider}
R.~Vedantam, C.~Lawrence~Zitnick, and D.~Parikh.
\newblock Cider: Consensus-based image description evaluation.
\newblock In \emph{Proceedings of the IEEE conference on computer vision and
  pattern recognition}, pages 4566--4575, 2015.

\bibitem[Wang et~al.(2021)Wang, Yu, Yu, Dai, Tsvetkov, and Cao]{wang2021simvlm}
Z.~Wang, J.~Yu, A.~W. Yu, Z.~Dai, Y.~Tsvetkov, and Y.~Cao.
\newblock Simvlm: Simple visual language model pretraining with weak
  supervision.
\newblock \emph{arXiv preprint arXiv:2108.10904}, 2021.

\bibitem[Zheng et~al.(2020)Zheng, Zhang, Li, Tang, Gao, and
  Zhou]{zheng2020structured3d}
J.~Zheng, J.~Zhang, J.~Li, R.~Tang, S.~Gao, and Z.~Zhou.
\newblock Structured3d: A large photo-realistic dataset for structured 3d
  modeling.
\newblock In \emph{Computer Vision--ECCV 2020: 16th European Conference,
  Glasgow, UK, August 23--28, 2020, Proceedings, Part IX 16}, pages 519--535.
  Springer, 2020.

\bibitem[Zhu et~al.(2017)Zhu, Park, Isola, and Efros]{zhu2017unpaired}
J.-Y. Zhu, T.~Park, P.~Isola, and A.~A. Efros.
\newblock Unpaired image-to-image translation using cycle-consistent
  adversarial networks.
\newblock In \emph{Proceedings of the IEEE international conference on computer
  vision}, pages 2223--2232, 2017.

\end{thebibliography}
\newpage
\appendix
\onecolumn
\section{Appendix.}

\subsection{Extra Implementation details}

\subsubsection{VQ-GAN details}
\label{sec:vq-gan-details}
\begin{table}[h]
\caption{Configuration and training hyperparameters for VQGAN.}
\begin{tabular}{l|l}
Config param &  Value  \\ \cline{1-2}
Perceptual loss weight & 0.05 \\
Adversarial loss weight & 0.01 \\
Codebook size & 8192 \\
Optimizer & Adam~\citep{kingma2014adam} \\
Discriminator learning rate & 1e-4 \\
Generator learning rate & 1e-4 \\
Optimizer momentum & $\beta_1$=0.9 $\beta_2$=0.99 \\
Batch Size & 256 \\
Learning rate schedule & Cosine \\ Decay~\citep{loshchilov2017decoupled}
Warmup steps & 10000 \\
Training steps & 500000 \\
\end{tabular}
\end{table}

VQGAN Architecture: Our VQGAN architecture is similar to the previous work~\citep{esser2021taming}. It consists of several
residual blocks, downsample (encoder) and upsample (decoder) blocks. The main difference is that we remove the non-local
block to make the encoder and decoder fully convolutional to support different image sizes. In the base VQGAN model, we
apply 2 residual blocks in each resolution and the base channel dimension is 128. For the finetuned decoder, we apply 4
residual blocks in each resolution and we also make the base channel dimension to be 256.

\subsubsection{Optimization details for \model}

\begin{table}[h]
\begin{tabular}{l|l}
Config param &  Value  \\ \cline{1-2}
Optimizer & AdamW \citep{loshchilov2017decoupled}  \\
Learning rate & 1e-4   \\
Warmup steps & 5e3 \\
Weight Decay & 1e-4 \\
Gradient Clip & 1.0 \\
\end{tabular}
\end{table}
\newpage

\subsubsection{Text to image generator details}
\label{sec:t2i-details}
\begin{table}[h]
\begin{tabular}{l|l}
Config param &  Value  \\ \cline{1-2}
Num Layers & 12  \\
Num Heads & 12   \\
Embedding hidden dim & 768 \\
MLP hidden dim & 3072 \\
Dropout & 0.1 \\
Codebook size & 8192 \\
Max Sample Length & 64 \\
Guidance Scale & 4 \\
Sample Temperature & 32.5 \\
Activation & GeLU \\
\end{tabular}
\end{table}

\subsubsection{Perceiver details}
\label{sec:perceiver-details}
\begin{table}[h]
\caption{Perceiver details}
\begin{tabular}{l|l}
Config param &  Value  \\ \cline{1-2}
Num Layers & 6  \\
Num Heads & 16   \\
Embedding hidden dim & 1024 \\
Activation & GeLU \\

\end{tabular}
\end{table}

\subsection{Finetuning details}
\label{sec:app-fine-tuning}

Following the pre-training stage, our model undergoes fine-tuning for various downstream tasks. The fine-tuning process employs the AdamW optimizer with similar Beta values as in pre-training. The learning rate was 1e-5.

To enhance generalization capabilities during fine-tuning, we utilize regularization methods such as Dropout (set to 0.1).

In accordance with established practices, we use the development split to determine optimal fine-tuning settings. The performance results are then reported based on the test split.

\subsection{Clustering details}
\label{sec:app-cluster}

To determine the optimal number of clusters for our analysis, we employed the Elbow Method in conjunction with the Within-Cluster Sum of Squares (WCSS) metric. We iteratively applied the K-means clustering algorithm while varying the number of clusters, calculating the WCSS for each iteration.  A clear `elbow' point was observed in the WCSS plot, indicating a substantial decrease in variance as the number of clusters increased up to 20. Beyond this point, further increases in the number of clusters yielded diminishing returns in terms of WCSS reduction.  Based on this analysis, we determined that 20 clusters provided a suitable balance between parsimony and capturing the underlying structure within our dataset.

To analyze the diversity of cluster compositions, we employed a co-clustering approach with the K-means algorithm on the concatenated GenPair, WebLI, and LTIP datasets.  For each of the resulting clusters, we calculated the normalized cluster sizes for each individual dataset. This allowed us to visualize the distribution among the datasets within each cluster, as illustrated in Figure \ref{fig:diversity}.

To calculate the concentration we performed a cumulative sum over the top-5 clusters to determine what percentage of the data points was present there. The idea is that, the higher the percentage  in the top 5 clusters, the less uniform is the distribution across all clusters, potentially indicating a lower level of diversity.
Here in Table \ref{tab:results-pure-synth-app} we also report top-3 and the entropy over the full 20 clusters. Both measures confirm a more uniform distribution of data across clusters for GenPair versus the WebLi and LTIP. Especially the entropy complement the other measures as it takes into account all clusters, even if they are smaller.

\begin{table*}[t]
\centering
\caption{Zero shot image captioning results when using synthetically generated caption and image embedding pairs. Concentration is calculated as the cumulative distribution on the top-5 clusters, a lower value represent higher diversity (see Appendix~\ref{sec:app-cluster} for more details).} 
\vspace{0.2cm}
\label{tab:results-pure-synth-app}
\resizebox{0.99\textwidth}{!}{\begin{tabular}{lcccccccccc}
\toprule
Real & Synth & \#Real Data & \#Synth Data & Concentration Top-3 ($\downarrow$) & Concentration Top-5 ($\downarrow$) & Entropy ($\uparrow$) & CIDEr-COCO ($\uparrow$) & Flickr-30($\uparrow$) & VQAV2 Acc. ($\uparrow$) & OKVQA Acc. ($\uparrow$)\\
\midrule
CCv2 & - & 10.1M & -  & - & - & - & 22.1 & 12.7 & 29.1 & 32.4\\
CCv2 & \dataset & 10.1M & 1M  & \textbf{40.5\%} & \textbf{57.7\%}  & \textbf{3.81} & \textbf{25.9} & \textbf{17.3} &\textbf{31.1} & \textbf{34.0}\\
CCv2+WebLI & -  & 10.1M+1M & -  & 58.2\% & 69.8\% & 3.43 & 24.4 & 14.9 & 30.6 & \textbf{33.9}\\
CCv2+LTIP &  - & 10.1M+1M & - & 76.5\% & 83.0\% & 2.92 &  23.4 & 13.8 & 30.3 & 32.9\\

\bottomrule
\end{tabular}}
\end{table*}

\subsection{Backbone Comparison}
We conducted an ablation study to compare the performance of different visual backbones. For this comparison, we kept the total amount of FLOPs (both pre-training and training) and (self)supervised training data constant across models.  We compared our VQ-GAN-based backbone against a contrastively pre-trained NFNet-F6 model \citep{brock2021high} using the same loss function as CLIP \citep{radford2021learning}.
As shown in Table~\ref{tab:backbone}, the primary factor influencing performance appears to be the number of parameters, as our model with a deeper VQ-Adapter (VIT-L+) achieved results comparable to the contrastively trained model.
In our study, we opted for the \model VIT-B backbone due to its compatibility with our hardware resources. This choice allowed us to avoid model sharding, thereby enabling faster experimental iterations.

\begin{table}[]
\centering
\caption{Backbone ablation study. In this table the total amount of FLOPS (pre-training and training) was held constant}
\vspace{0.1cm}
\label{tab:backbone}
\resizebox{0.7\textwidth}{!}{\begin{tabular}{lccccc}
\toprule
Model & real data & \#params $\uparrow$ & MSCOCO CIDEr $\uparrow$ &
VAQV2 Accuracy $\uparrow$\\
\midrule
\model \space VIT-B adapt. & 690.1 & 632M & 35.3 & 37.9\\
\model \space VIT-L+ adapt. & 690.1 & 890M & 40.9 & 42.8\\
NFNet-F6 (CLIP) & 690.1 & 920M & 41.3 & 43.4 \\
\bottomrule
\end{tabular}}
\end{table}

\subsection{Image Generator Qualitatives}
\begin{figure*}[ht]
  \centering
    \includegraphics[width=.6\linewidth]{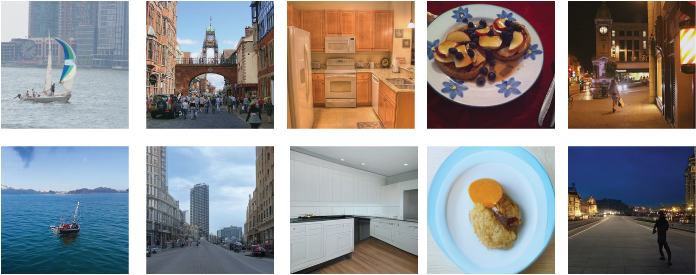}
    \caption{Qualitatives showing images selected from the validation set of MS-COCO~\citep{chen2015microsoft} (top row) and their synthetic versions generated by our text-to-image generator given the ground truth captions (bottom). The training of the image generator has been done on Conceptual Caption v2~\citep{changpinyo2021conceptual}.}
    \label{fig:gen-images}
\end{figure*}

\end{document}